\documentclass[preprint,12pt]{elsarticle}
\usepackage{graphicx}
\usepackage{amsmath,amssymb}
\usepackage{caption,subcaption}
\usepackage{multirow}
\usepackage{color}
\usepackage{amssymb}
\newcommand{\ignore}[1]{}
\usepackage{graphicx}
\usepackage{hyperref}
\usepackage{url}
\usepackage{amsmath,amssymb}
\usepackage{hyperref}
\usepackage{algorithmic}
\usepackage{algorithm}
\usepackage{color}
\usepackage{threeparttable}
\usepackage{multirow}
\usepackage{makecell}
\usepackage{amsmath}
\usepackage{booktabs}
\usepackage{caption}
\begin{document}
\begin{frontmatter}
\title{Title\tnoteref{label1}}
\author{Kaihua Zhang}
\ead{zhkhua@gmail.com}
\author{Xuejun Li}
\author{Qingshan Liu}
\cortext[cor1]{Corresponding author}
\address{Jiangsu Key Laboratory of Big Data Analysis Technology (B-DAT), Nanjing University of Information Science and Technology. }
\title{Unsupervised Video Segmentation via Spatio-Temporally Nonlocal Appearance Learning }
\begin{abstract}

Video object segmentation is challenging due to the factors like rapidly fast motion, cluttered backgrounds, arbitrary object appearance variation and shape deformation.
Most existing methods only explore appearance information between two consecutive frames, which do not make full use of the usefully long-term nonlocal information that is helpful to make the learned appearance stable, and hence they tend to fail when the targets suffer from large viewpoint changes and significant non-rigid deformations.
%
In this paper, we propose a simple yet effective approach to mine the long-term sptatio-temporally nonlocal appearance information for unsupervised video segmentation.
The motivation of our algorithm comes from the spatio-temporal nonlocality of the region appearance reoccurrence in a video. Specifically, we first generate a set of superpixels to represent the foreground and background, and then update the appearance of each superpixel with its long-term sptatio-temporally nonlocal counterparts generated by the approximate nearest neighbor search method with the efficient KD-tree algorithm. Then, with the updated appearances, we formulate a spatio-temporal graphical model comprised of the superpixel label consistency potentials. Finally, we generate the segmentation by optimizing the graphical model via iteratively updating the appearance model and estimating the labels.
Extensive evaluations on the SegTrack and Youtube-Objects datasets demonstrate the effectiveness of the proposed method, which performs favorably against some state-of-art methods.
\end{abstract}
\end{frontmatter}

\section{Introduction}
Video object segmentation is a task of separating the moving foreground object consistently from the complex background in unconstrained video sequences.
Although much progress has been made in the past decades, it remains a challenging task due to the factors such as fast motion, cluttered backgrounds, arbitrary object appearance variation and shape deformation, to name a few.

One key step to deal with this problem is to maintain both spatial and temporal consistency across the whole video, based on which numerous methods have been proposed, which can be generally categorized into two categories: supervised segmentation and unsupervised segmentation.
The supervised video segmentation requires a user to manually annotate some frames, which guide the segmentation of other frames across all frames.
Most supervised methods are graph-based~\cite{marki2016bilateral,wang2012probabilistic,price2009livecut,bai2009video}, which usually include a unary term comprised of foreground appearance, motions or locations and a pairwise term that encodes spatial and temporal smoothness to propagate the user$'$s annotations to all other frames.
Moreover, the optical flows are usually adopted to deliver information among frames, but it is prone to failure because of the inaccurately estimated optical flows.
To address this issue, some methods based on tracking~\cite{wen2015jots,tsai2012motion,brendel2009video,li2013video,wang2011superpixel} have been proposed, which first label the position of the object in the first frame, and then enforce the temporal consistency in the video by tracking pixels, superpixels or object proposals.
However, most of those approaches only consider the pixels or superpixels that are generated independently in each frame without exploiting the ones from the long-term spatio-temporal regions, which are helpful to learn a robust appearance model.
In contrast to the supervised segmentation, a variety of unsupervised video segmentation algorithms have been proposed in recent years~\cite{brox2010object,taylor2015causal,papazoglou2013fast,shankar2015video,ochs2012higher,lee2011key}, which are fully automatic without any manual interventions.
\cite{brox2010object,ochs2012higher} are based on clustering point trackers, which can integrate information of a whole video shot to detect a separately moving object, among which~\cite{brox2010object} explores point trajectories as well as motion cues over a large time window, which is less susceptible to the short-term variations that may hinder separating different objects.~\cite{lee2011key,li2013video} are based on object proposals, which utilize appearance features to calculate the foreground likelihood and match partial shapes by a localization prior. Besides, some other segmentation methods have been proposed, which consider occlusion cues~\cite{taylor2015causal} and motion characteristics~\cite{papazoglou2013fast} to hypothesize the foreground locations.

\begin{figure*}[t]
\begin{center}
\begin{tabular}{c}
\includegraphics[width=1\textwidth]{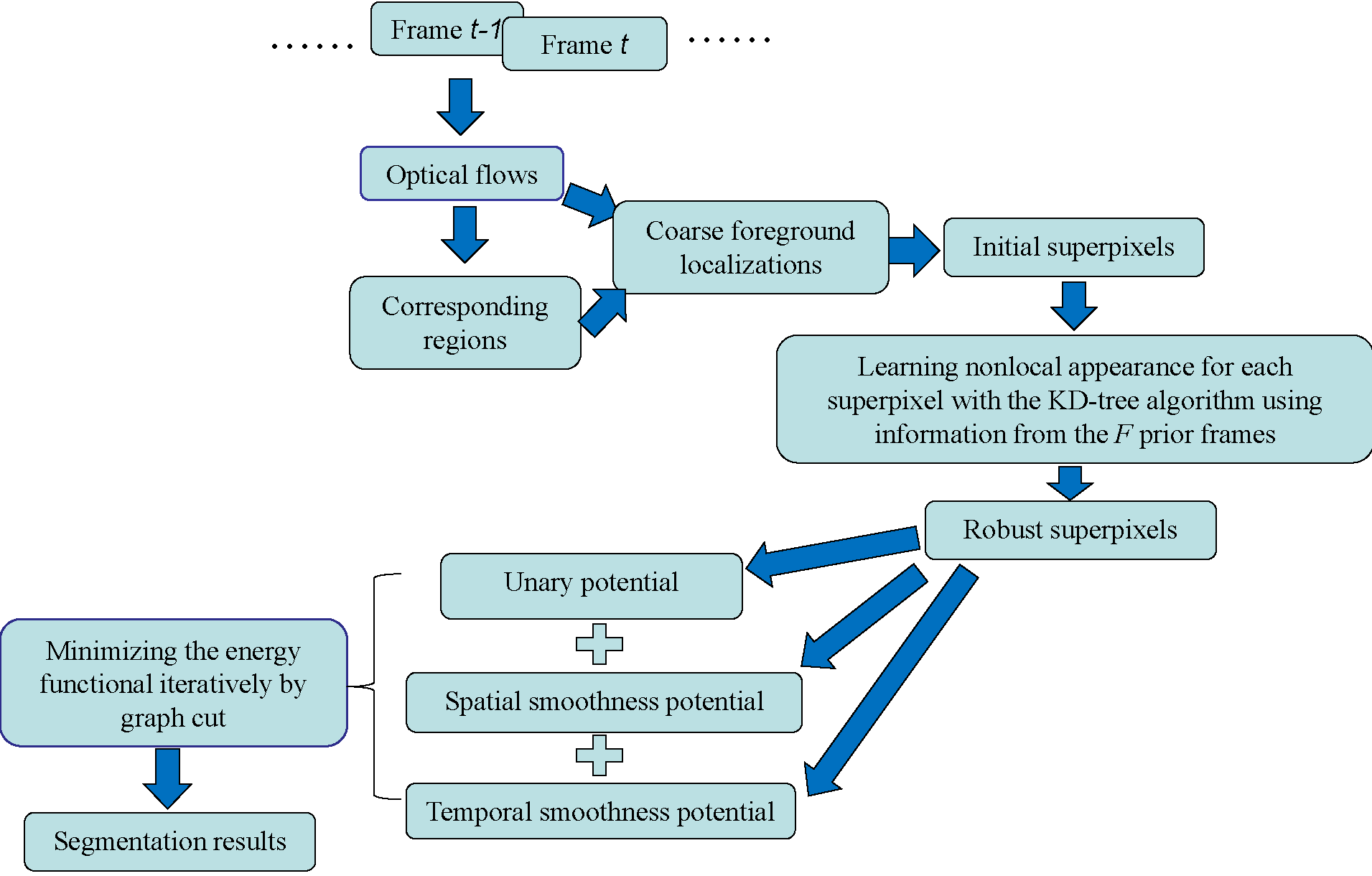}
\end{tabular}
\end{center}
\caption{Flow chart of our method}
\label{fig:flow}
\end{figure*}
In this paper, we propose a fully automatic video object segmentation algorithm without the help of any extra knowledge about the object position, appearance or scale. Figure~\ref{fig:flow} shows the flow chart of our method. First, we utilize the optical flow information to obtain a rough object position that ensures the frame-to-frame segmentation consistency.
Specifically, we employ the method of~\cite{papazoglou2013fast}, which can produce a rough motion boundary in pairs of adjacent frames and then get an efficiently initial foreground estimation. Here, the only requirement for the object is to move differently from its surrounding background in some frames of the video.
Moreover, in order to reduce the noises introduced by appearance learning, we explore the information of superpixels from the long-term spatio-temporally nonlocal regions to learn a robust appearance model, which is integrated into a spatio-temporal graphical model.
Finally, as GrabCut~\cite{rother2004grabcut}, the graphical model is iteratively solved by refining the foreground-background labeling and updating the foreground-background  appearance models. We evaluate the proposed algorithm on two challenging datasets, i.e. SegTrack~\cite{tsai2012motion} and Youtube-Objects~\cite{prest2012learning}, and show favorable results against some state-of-art methods.
%
\section{Methodology}
In contrast to the supervised methods~\cite{wen2015jots,marki2016bilateral}, our method does not need to give the initialization in the first frame as a prior. Before assigning each pixel a label, in order to reduce the computational complexity and the background noise, we first use the method introduced in Section~\ref{background} to obtain the coarse object location mask in each frame. Then, we use the TurboPixel algorithm~\cite{levinshtein2009turbopixels} to oversegment the whole video sequence into a set of superpixels, which are used to generate the initially hypothesized models. Then, the appearances of the superpixels are updated by their spatio-temporally nonlocal counterparts from several distant frames. Finally, with these updated superpixels, we design a spatial-temporal graphical model to assign each superpixel with a foreground or background label.
\subsection{Foreground Localization for Coarse Segmentation}
\label{background}
As in~\cite{papazoglou2013fast}, we first coarsely localize the foreground with motion information, in which the rough motion boundaries can be estimated by integrating both the gradient and direction information of the optical flows.
Let $\vec{F}_i$ denote the optical flow vector at pixel $i$, $B_i^m\in[0,1]$ be the strength of the motion boundary at pixel $i$, and $B_i^\theta\in[0,1]$ be the difference of the directions between the motion of pixel ${i}$ and its neighboring pixels in set $\mathcal{N}_i$. Then the probability of the motion boundary is estimated as

\begin{equation}
P_{i}=\begin{cases}B_{i}^m
 & \text{ if } B_{i}^m>\rho, \\ B_{i}^m\cdot B_{i}^\theta
 & \text{ if } B_{i}^m\leq \rho,
\end{cases}
\end{equation}
where $B_{i}^m$ is defined as
\begin{equation}
B_{i}^m=1-\exp(-\lambda^m||\nabla \vec{F_i}|| ),
\end{equation}
where $\lambda^m$ is a parameter that controls the steepness of the
function, and $B_i^\theta$ is defined as

\begin{equation}
B_i^\theta =1-\exp(-\lambda^\theta \underset{j\in \mathcal{N}_i}{\max}(\theta ^2_{i,j})),
\end{equation}
where $\theta_{i,j}$ denotes the angle between $\vec{F_i}$ and $\vec{F_j}$, and $\rho$ is a threshold which is set to 0.5 in our experiments.
Finally, we threshold $P_i$ at 0.5 to produce a binary motion boundary labeling.

After getting the rough motion boundaries, an inside-outside map based on the point-in-polygon problem~\cite{hughes2014computer} is then exploited, which uses the integral images~\cite{wang2012probabilistic} to generate a rough object mask. By shooting 8 rays spaced by 45 degrees, and if a ray intersects the boundary of the polygon an odd number of times, the start point of the ray is inside the polygon. Finally, the majority voting rule is adopted to decide which pixels are inside, resulting is the inside-outside map $\mathcal{M}$.

The estimated optical flows may be inaccurate when the foreground object moves abruptly, thereby leading the method unable to extract the exact object location or shape in certain frames. Notwithstanding, the inside-outside map $\mathcal{M}$ ensures that most of pixels within the object can be covered.

\subsection{Spatio-Temporal Graphical Model for Refining Segmentation}
In this section, we employ a spatio-temporal graphical structure, which allows to propagate the appearance information from some spatio-temporally distant regions in the video. Different from~\cite{li2013video} whose neighborhoods are nonlocal only in space, our method explores the long-term nonlocal appearance information both in space and in time. By taking into account the long-term cues (visual similarities across several frames), the background noise can be effectively reduced.

%
Let $\mathcal{S}=\left \{\mathcal{S}^{1},\mathcal{S}^{2},\cdots ,\mathcal{S}^{k}\right \}$ represent the set of all superpixels in $k$ frames in a video sequence, where $\mathcal{S}^t=\left \{ \mathcal{S}_1^t,\mathcal{S}_2^t,\cdots ,\mathcal{S}_n^t \right \}$ is the set of all superpixels in the $t$-th frame, and $\mathcal{S}_i^t=(\mathcal{A}_i^t,\mathcal{P}_i^t,l_i^t)$ denotes the $i$-th superpixel in the $t$-th frame.
$\mathcal{A}_i^t$ is an appearance model constructed with the average HSV and RGB color features of the corresponding pixels inside the superpixel $i$.
$\mathcal{P}_i^t$ is the center location of the superpixel $i$, and $\mathit{l_i^t}\in \left \{ 0,1 \right \}$ is the label indicating that the superpixel $i$ belongs to background or foreground respectively. Similar to~\cite{papazoglou2013fast}, the segmentation is formulated by evaluating a labeling via minimizing the energy functional
\begin{equation}
\label{energy}
E(\{l_i^t\}_{i,t})=\sum_{i,t} U_i^t(\mathcal{S}_i^t)+\gamma _1 \sum_{\left (i,j\right )\in \mathcal{N}_s,t}V_{ij}^t\left(\mathcal{S}_i^t,\mathcal{S}_j^t\right )+\gamma _2 \sum_{\left (i,j\right )\in \mathcal{N}_t,t}W_{ij}^t\left (\mathcal{S}_i^t ,\mathcal{S}_j^{t-1} \right ),
\end{equation}
where $U_i^t(\mathcal{S}_i^t)$ is a unary potential for labeling the $i$-th superpixel in the $t$-th frame as foreground or background defined as
\begin{equation}
\label{eq:unary}
U_i^t(\mathcal{S}_i^t) = A^t(\mathcal{S}_i^t)+L^t(\mathcal{S}_i^t,\mathcal{M}^{t-1}),
\end{equation}
where $A^t(\cdot)$ is the color score function on the superpixel $\mathcal{S}_i^t$ constructed by a Gaussian Mixture Model (GMM), and $L^t(\cdot)$ is the location score function designed by the Euclidean distance transform of the mask of the inside-outside map $\mathcal{M}^{t-1}$ in the $(t-1)$-th frame.
$V_i^t(\cdot)$ and $W_i^t(\cdot)$ are two pairwise potentials that measure spatial and temporal smoothness with weights $\gamma_1$ and $\gamma_2$ respectively. $\mathcal{N}_s$ is the set of the spatially adjacent neighbors of the superpixels in the same frame while $\mathcal{N}_t$ denotes the temporally adjacent neighbors of the superpixels between two consecutive frames. As GrabCut~\cite{rother2004grabcut}, the solution of minimizing $E(\{l_i^t\}_{i,t})$ can be achieved by iteratively updating the appearance model and estimating the labels of foreground and background.
\subsubsection{Spatio-Temporally Nonlocal Appearance Learning}
In~\cite{papazoglou2013fast}, the superpixels are generated independently in each frame, which are connected by optical flows to enhance the motion consistency between two consecutive frames. However, the inaccurately estimated optical flows may degrade the reliability of the connections. Moreover, the connections are only based on the consecutive frames, which do not make full use of the long-term spatio-temporal information that is helpful to reduce the noise introduced by significant appearance variations over time. To address these issues, we propose a simple yet effective approach that mines the long-term spatio-temporally coherent superpixels to learn a robust appearance model directly from the feature space of the superpixels without resorting to optical flows. Below, we itemize the details.

\begin{figure*}[h]
\begin{center}
\begin{tabular}{c}
\includegraphics[width=.81\textwidth]{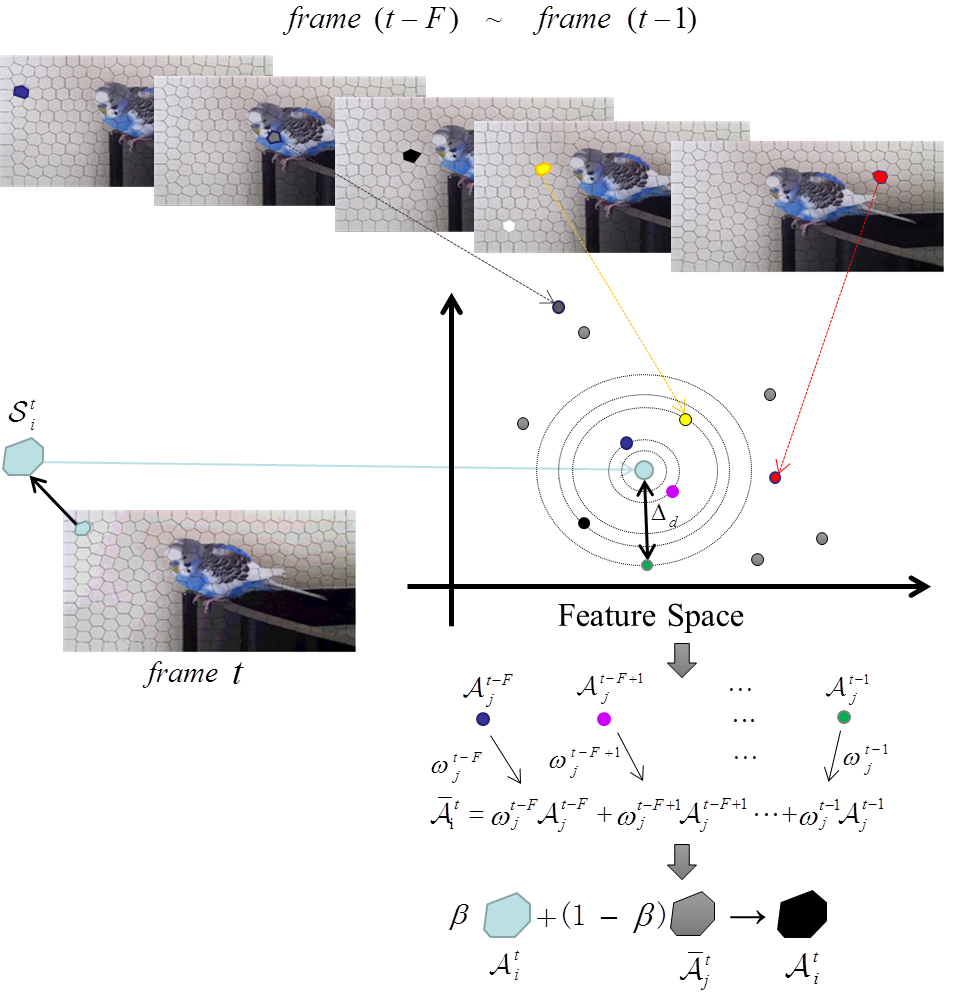}
\end{tabular}
\end{center}
\caption{Illustration of how to mine the nonlocal appearance information from the long-term spatio-temporally coherent superpixels.}
\label{fig:overview}
\end{figure*}
Figure~\ref{fig:overview} shows the procedure of how to capture the spatio-temporally nonlocal appearance information. For superpixel $\mathcal{S}_i^t$, we search its $F$ spatio-temporally coherent counterparts $\{\mathcal{S}_j^{t^\prime}\}_{t^\prime=t-F}^{t-1}$ from the formerly consecutive $F$ frames via the approximate nearest neighbour search method that utilizes an efficient KD-tree search algorithm provided by VL-feat~\cite{vedaldi2010vlfeat}.
For the purpose of efficiency, we restrict the nearest neighbour search space in $F=5$ frames.
Motions are limited in such a small time interval, which increases the chance of matching foreground with foreground, and background with background.
We linearly combine these extracted superpixel appearances as
\begin{equation}
\bar{\mathcal{A}_i^t}=\sum_{{t}^{\prime}=t-F}^{t-1}\omega_j^{t^\prime}\mathcal{A}_j^{t^\prime},
\end{equation}
where the weight $\omega_j^{t^\prime}$ is defined as
\begin{equation}
\omega_j^{t^\prime}=\frac{\exp(-\Delta_d(\mathcal{A}_i^t,\mathcal{A}_j^{t^\prime}))}{\sum_{t^\prime=t-F}^{t-1}\exp(-\Delta_d(\mathcal{A}_i^t,\mathcal{A}_j^{t^\prime}))},
\end{equation}
where $\Delta_d(\cdot,\cdot)$ denotes the Euclidean distance between two appearance feature vectors.
Finally, we update the superpixel appearance $\mathcal{A}_i^t$ by integrating the long-term spatio-temporal appearance $\bar{\mathcal{A}_i^t}$
\begin{equation}
\label{eq:updateA}
 \mathcal{A}_i^t=\beta\mathcal{A}_i^t+(1-\beta)\bar{\mathcal{A}_i^t},
\end{equation}
which is put into the energy functional $E(\{l_i^t\}_{i,t})$ in~(\ref{energy}).
\subsubsection{Spatial Smoothness Potential}
The spatial smoothness term $V_{ij}^t$ in~(\ref{energy}) models the interactions between two neighboring superpixels $i$ and $j$ in the $t$-th frame.
Specifically, if a superpixel is labeled as foreground, we expect that the superpixel has a small energy with respect to the foreground in two aspects: First, intuitively, two adjacent superpixels with similar appearances are more likely to belong to the same segmented region with a small energy.
%
Therefore, restricting the spatial neighbors with similar appearances is able to reduce the chance of confusing foreground and background regions.
Second, the adjacent superpixels with a small spatial distance would be more likely to be assigned to the same label with a small energy.
Therefore, the appearance and location cues provide effective information to distinguish the object superpixels from their background ones, which are explored to design the spatial smoothness term as
\begin{equation}
V_{ij}^t(\mathcal{S}_i^t,\mathcal{S}_j^t)=\mathbb{I}(l_i^t\neq l_j^t)\cdot\exp(-\alpha\cdot\Delta_d(\mathcal{A}_i^t,\mathcal{A}_j^t))\cdot (\Delta_d(\mathcal{P}_i^t,\mathcal{P}_j^t))^{-1}
\end{equation}
where $\mathbb{I}(l_i^t \neq l_j^t)$ returns 1 if its input argument is true and 0 otherwise, and $\Delta_d(\mathcal{P}_i^t,\mathcal{P}_j^t)$ is the Euclidean distance between the centers of the two adjacent superpixels $i$ and $j$.
\subsubsection{Temporal Smoothness Potential}
The temporal smoothness potential $W_{ij}^t$ in~(\ref{energy}) measures the interactions of pairs of temporally connected superpixels between two consecutive frames. In~\cite{papazoglou2013fast}, only the superpixel appearances in two consecutive frames are explored to measure the appearance similarities, which are inaccurate when the target appearances vary significantly. To address this issue, we employ the updated superpixel appearances $\mathcal{A}_i^t$ and $\mathcal{A}_i^{t-1}$ via (\ref{eq:updateA}) that considers the spatio-temporally nonlocal appearance information to build the temporal smoothness potential as
\begin{equation}
W_{ij}^t(\mathcal{S}_i^t,\mathcal{S}_j^{t-1})=\mathbb{I}(l_i^t\neq l_j^t)\cdot\exp(-\alpha\cdot\Delta_d(\mathcal{A}_i^t,\mathcal{A}_j^{t-1}))\cdot\psi(\mathcal{A}_i^t,\mathcal{A}_j^{t-1}),
\end{equation}
where $\psi(\mathcal{A}_i^t,\mathcal{A}_j^{t-1})$ is the percentage of pixels connected by the optical flows from superpixel $\mathcal{S}_j^{t-1}$ to superpixel $\mathcal{S}_i^{t}$.

\section{Experimental Results}
\subsection{Implementation Details}
We evaluate the proposed method on two widely used video segmentation benchmark datasets, namely SegTrack dataset~\cite{tsai2012motion} and YouTube-Objects dataset~\cite{prest2012learning}.
For fair comparison, as in~\cite{papazoglou2013fast}, the results in terms of the average pixel error per frame are reported on the SegTrack dataset and the results with the intersection-over-union overlap metric are reported on the YouTube-Objects dataset.
%

We use Turbopixels~\cite{levinshtein2009turbopixels} to generate a set of superpixels in each frame. Likewise, SLIC~\cite{achanta2012slic} can also be adopted with much faster performance. However, we have found that using the SLIC algorithm resulted in a little decrease in segmentation accuracy in our experiments.
Each sequence in the SegTrack dataset generates about 50$\sim$100 superpixels per frame and for the sequences in the YouTube-Objects dataset, we generate up to 1500 superpixels per frame.
Table~\ref{tab:segtrack} and Table~\ref{tab:youtube} report the quantitatively evaluated results against several state-of-art methods~\cite{jain2014supervoxel,vijayanarasimhan2012active,godec2013hough,papazoglou2013fast,ochs2014segmentation,wen2015jots,tsai2012motion,cai2014robust,ochs2012higher,li2013video,lee2011key},
in which the top two ranked methods are highlighted in red and blue, respectively. Furthermore, Figures~\ref{fig:segtrack1},~\ref{fig:youtube1} and~\ref{fig:youtube2} present some qualitative segmentation results generated by our method.
\begin{table}[t]
\begin{center}
\begin{tabular}{|c|c|c|c|c|c|c|c|c|c|}
\hline
Sequence & \cite{wen2015jots} & \cite{tsai2012motion} & \cite{cai2014robust} & \cite{jain2014supervoxel} & \cite{ochs2012higher} & \cite{papazoglou2013fast} & \cite{li2013video} &\cite{lee2011key} & Ours\\
\hline
Unsupervised & $\times$ & $\times$ & $\times$ & $\times$ & $\surd$ & $\surd$ & $\surd$ & $\surd$ & $\surd$\\
\hline
Birdfall & {\color{red}163} & 252 & 481 &{\color{blue} 189} & 468 & 217 & 242 & 288 &211\\
Cheetah& {\color{red}806}& 1142 & 2825 & 1170 & 1175 & 890 & 1156 & 905 &{\color{blue}813}\\
Girl & 1904 & {\color{red}1304} & 7790 & 2883 & 5683 & 3859 & {\color{blue}1564} & 1785 &2269\\
Monkeydog& 342 & 563 & 5361 & 333 & 1434 & {\color{red}284 }& 483 & 521 &{\color{blue}308}\\
Parachute& 275 & 235 & 3105 & {\color{blue}228} & 1595 & 855 & 328 & {\color{red}201} &353\\
\hline
\end{tabular}
\caption{Average pixel errors per frame (The lower the better) for some representative state-of-art methods on the SegTrack dataset.}
\label{tab:segtrack}
\end{center}
\end{table}
\subsection{Results on the SegTrack Dataset}
The SegTrack dataset~\cite{tsai2012motion} was designed to evaluate object segmentation in videos which consists of 6 challenging video sequences (``Birdfall'', ``Cheetah'', ``Girl'', ``MonkeyDog'', ``Parachute'' and ``Penguin'') with pixel-level human annotated segmentation results of the foreground objects in every frame.
Videos in this dataset contain 21$\sim$71 frames each with several challenging factors like color overlap in objects, large inter-frame motion, shape and appearance changes and motion blur. The standard evaluation metric is the average pixel error that is defined as the average number of mislabeled pixels over all frames per video~\cite{tsai2012motion}.

%
Table~\ref{tab:segtrack} shows the quantitative results in terms of the average pixel error per frame of the proposed algorithm and other representative state-of-art methods including tracking and graph-based approaches~\cite{wen2015jots,tsai2012motion,jain2014supervoxel,ochs2012higher,papazoglou2013fast,li2013video,lee2011key}. Note that our method is fully automatic while some compared methods are supervised that are marked in the table.
Overall, the proposed algorithm achieves favorable results in most sequences especially for those with non-rigid objects. Our method outperforms~\cite{cai2014robust,ochs2012higher} in all videos, and outperforms all other algorithms except for~\cite{wen2015jots} in 3$\sim$4 video sequences, including those supervised methods~\cite{tsai2012motion,cai2014robust,jain2014supervoxel}.
Moreover, our algorithm achieves a comparable result with~\cite{wen2015jots} with a much easier implementation as our method is fully automatic while~\cite{wen2015jots} needs to give the manually selected object regions in the first frame.
Special notice should be taken on the substantial gains of our method on the challenging ``Monkeydog" and ``Cheetah" sequences, which suffer from large deformations due to fast motions and complex cluttered backgrounds.
The proposed method achieves the second best results on these two sequences among all approaches by a narrow margin to the best one.

\begin{figure*}[!htb]
\begin{center}
\begin{tabular}{c}
\includegraphics[width=1\linewidth]{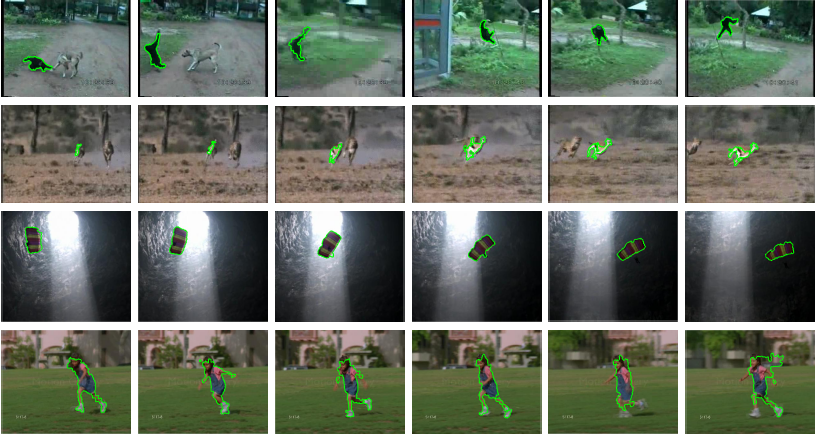}
\end{tabular}
\end{center}
\caption{Example results for segmentation on four sequences from SegTrack dataset. Top to bottom: ``Monkeydog", ``Cheetah", ``Parachute" and ``Girl"  sequences}
\label{fig:segtrack1}
\end{figure*}
Figure~\ref{fig:segtrack1} shows some example results from ``Monkeydog", ``Cheetah", ``Parachute" and ``Girl" sequences. Our method successfully propagates the foreground of ``Monkeydog" sequence despite it suffers from considerable motions and deformations, and so does the ``Cheetah" sequence that has very similar appearances between the foreground and background. On the contrary, our method achieves a little weak performance on the ``Girl" and ``Parachute" sequences, which are due to the severe motion blur and the adopted superpixel-based representations. Our method totally depends on superpixel-level appearances, which may not perform well on some sequences with complex target objects because the superpixels cannot preserve object boundaries well. Also, when encountering serious motion blur, for example, when the target object moves quickly or suffers from low resolution (See the ``Girl" sequence), it is difficult for our method to get an accurate matching between the consecutive frames because the optical flow links that our propagation are based on may suffer from severe errors and drift.

\begin{table}[t]
\begin{center}
\begin{tabular}{|c|c|c|c|c|c|c|}
\hline
Category & \cite{jain2014supervoxel} & \cite{vijayanarasimhan2012active} & \cite{godec2013hough} & \cite{papazoglou2013fast} & \cite{ochs2014segmentation} &Ours\\
\hline
Unsupervised & $\times$ & $\times$ & $\times$ & $\surd$ & $\surd$ & $\surd$ \\
\hline
aeroplane & {\color{red} 86.3 } &{\color{blue}79.9} & 73.6 & 70.9 & 13.7 & 77.6\\
bird& {\color{red}81.0} & 78.4 & 56.1 & 70.6 & 12.2 &{\color{blue}78.9}\\
boat& {\color{red}68.6} & 60.1 & 57.8 & 42.5 & 10.8 &{\color{blue}60.4}\\
car& {\color{blue}69.4} & 64.4 & 33.9 & 65.2 & 23.7 &{\color{red}73.0}\\
cat& {\color{blue}58.9} & 50.4 & 30.5 & 52.1 & 18.6 &{\color{red}63.8}\\
cow& {\color{red}68.6} & 65.7 & 41.8 & 44.5 & 16.3 &{\color{blue}65.9}\\
dog& 61.8 & 54.2 & 36.8 & {\color{blue}65.3} & 18.0 &{\color{red}65.6}\\
horse& {\color{blue}54.0} & 50.8 & 44.3 & 53.5 & 11.5 &{\color{red}54.2}\\
motorbike& {\color{red}60.9} & {\color{blue}58.3} & 48.9 & 44.2 & 10.6 & 53.8\\
train& {\color{red}66.3} & {\color{blue}62.4} & 39.2 & 29.6 & 19.6 & 35.9\\
\hline
Mean& {\color{red}67.6} & 62.5 & 46.3 & 53.8 & 15.5 & {\color{blue}62.9}\\
\hline
\end{tabular}
\caption{Quantitative results in terms of the intersection-over-union overlap metric on the Youtube-Objects dataset (The higher the better).}\label{tab:youtube}
\end{center}
\end{table}
\subsection{Results on the Youtube-Objects Dataset}
The Youtube-Objects~\cite{prest2012learning} is a large dataset that contains 1407 video shots with 10 object categories from the internet, and the length of each sequence can be up to 400 frames.
Videos in this dataset are completely unconstrained with large camera motion, complex background, rapid object moving, large scale viewpoint changes and non-rigid deformation, etc, which make it very challenging.
We use a subset of the Youtube-Objects dataset defined by~\cite{tang2013discriminative}, which includes 126 videos with more than 20000 frames with provided segmentation ground truth.
However, the ground truth provided by~\cite{tang2013discriminative} is approximate because the annotators marked the superpixels computed by~\cite{grundmann2010efficient}, but not the individual pixels. So, we employ the fine-grained pixel-level annotations of the target objects in every 10 frames provided by~\cite{jain2014supervoxel}.

Table~\ref{tab:youtube} shows the quantitative results in terms of the overlap accuracy of the proposed algorithm and other representative state-of-art methods. For the tracking or foreground propagation based algorithms~\cite{jain2014supervoxel,vijayanarasimhan2012active,godec2013hough}, the ground-truth annotations of the first frames are used to initialize the propagated segmentation masks.
%
Generally speaking, our method performs well in terms of the overlap ratio, especially on 7 out of 10 categories. As shown by Table~\ref{tab:youtube}, our method substantially outperforms the unsupervised video segmentation method~\cite{papazoglou2013fast} in terms of the mean overlap ratio by more than $9\%$ from 53.8 to 62.9.
Moreover, our method outperforms the unsupervised method~\cite{ochs2014segmentation} by a large margin with more than $47\%$ of the mean overlap ratio.
In addition, compared with those supervised algorithms~\cite{jain2014supervoxel,vijayanarasimhan2012active,godec2013hough}, the proposed method achieves the best performance on 4 categories and the second best performance on another 3 categories.
Considering that our method does not resort to any extra information from the ground truth, the results are satisfying.
Especially, the proposed algorithm performs well on the fast moving objects such as ``car" and ``dog" sequences as the errors introduced by the inaccurately estimated optical flows can be reduced by taking into account the long-term appearance information.
Recently, the supervised method~\cite{jain2014supervoxel} also demonstrated a better performance on the same sequences because it explores the long-term appearance and motion information from the supervoxels to enhance the temporal connections. Our method achieves a comparable result with~\cite{jain2014supervoxel} but without depending on any manually input information.

\begin{figure*}[t]
\begin{center}
\begin{tabular}{c}
\includegraphics[width=1\linewidth]{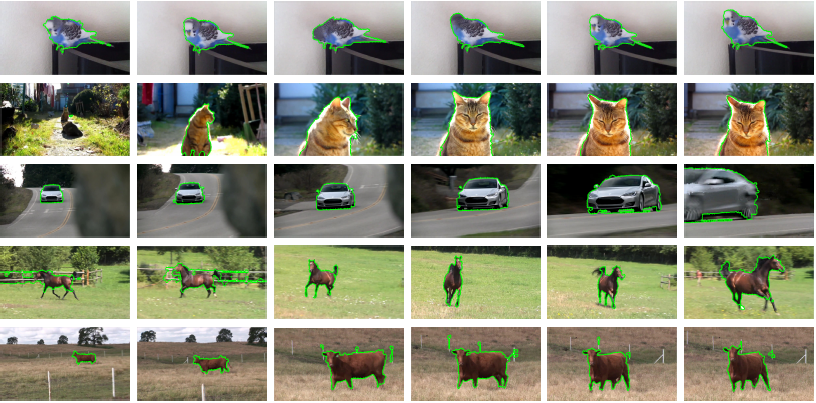}
\end{tabular}
\end{center}
\caption{Example results for segmentation on the video sequences from the YouTube-Objects dataset. Top to bottom: ``bird'',``cat'', ``car'', ``horse'' and ``cow''}
\label{fig:youtube1}
\end{figure*}

Figure~\ref{fig:youtube1} shows some qualitative results for five sequences ``bird'',``cat'', ``car'', ``horse'' and ``cow'' with non-rigid targets. Our method performs well even in the case that there exist significant object or camera motions. Furthermore, as we take the long-term appearance information into consideration, the segmentation results delineate the boundaries of the targets well especially for the non-rigid objects.

\begin{figure*}[!htb]
\begin{center}
\begin{tabular}{c}
\includegraphics[width=1\linewidth]{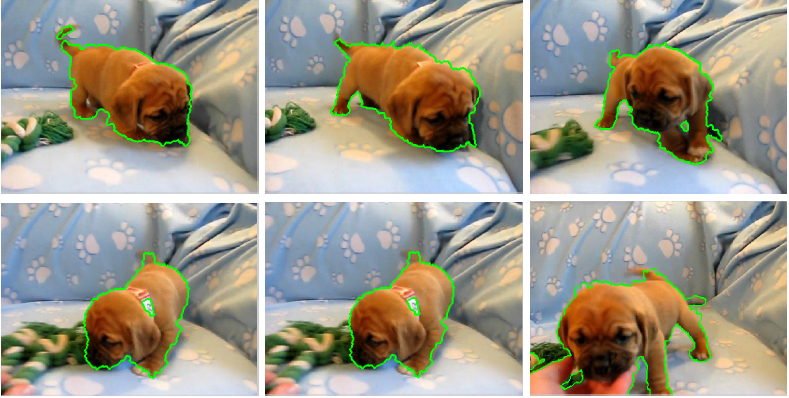}
\end{tabular}
\end{center}
\caption{Example results for segmentation on the video sequence ``dog'' from the YouTube-Objects dataset.}
\label{fig:youtube2}
\end{figure*}

Although our method has already achieved relatively satisfying segmentation results for most sequences, however, it also meets some problems in certain video sequences such as the ``dog'' sequence in Figure \ref{fig:youtube2}. Since our method does not designate the target region in the first frame, it searches the object totally based on the optical flows, and hence all the regions with apparent movement will be indicated as the target objects. As shown in Figure \ref{fig:youtube2}, our method also separates the region of the hand out because there is no information provided to tell that the target is just the dog. Therefore, our method may increase the errors especially for those sequences with multiple moving regions or partial target objects.

\section{Conclusion}
In the paper, we have presented a novel unsupervised video segmentation approach that effectively explores the long-term spatio-temporally nonlocal appearance information. Specifically, we updated the appearance of each superpixel by its spatio-temporally nonlocal neighbor counterparts extracted with the nearest neighbor search method implemented by the efficient KD-tree algorithm. Then, we integrated these updated appearances into a spatio-temporal graphical model, via optimizing which we generated the final segmentation. We have analyzed the impact of this updated appearance information on the SegTrack and Youtube-Objects datasets and found that the long-term appearances contribute a lot to improve the algorithm$^\prime$s robustness. Particularly, our approach deals well with the challenging factors such as large viewpoint changes and non-rigid deformation. Extensive evaluations on the two benchmark datasets demonstrated that our method performed favorably against some representative state-of-art video segmentation methods.
\section*{References}
\bibliographystyle{plain}
\bibliography{referance}
\end{document}